\documentclass[11pt]{article}
\usepackage{acl}  
\usepackage{times}
\usepackage{latexsym}
\usepackage[T1]{fontenc}
\usepackage[utf8]{inputenc}
\usepackage{microtype}
\usepackage{inconsolata}
\usepackage{graphicx}
\usepackage{hyperref}
\usepackage{amsmath}
\usepackage{amssymb}
\usepackage{pifont}
\usepackage{booktabs}
\newcommand{\cmark}{\ding{51}}
\newcommand{\xmark}{\ding{55}}

\newif\ifanonymize
\anonymizefalse

\title{The Hidden States Cookbook: A Large-Scale Ablation Study for Noise-Robust Conversational Intent Classification in Industry}

\author{
  \ifanonymize
    \textbf{Anonymous authors}
  \else
    \textbf{Bogdan Bogachov\textsuperscript{1,2}},
    \textbf{Nikita Letov\textsuperscript{2}},
    \textbf{Yaoyao Fiona Zhao\textsuperscript{1}}
  \fi
  \\
  \\
  \ifanonymize
    \small{Anonymous institutions}
  \else
    \textsuperscript{1}Department of Mechanical Engineering, McGill University, Montréal, Québec, Canada\\
    \textsuperscript{2}Axya Inc., Montréal, Québec, Canada\\
    \\
    \small{
      \texttt{bogdan.bogachov@mail.mcgill.ca},  \texttt{nikita@axya.co}, \texttt{yaoyao.zhao@mcgill.ca}
    }
  \fi
}

\begin{document}
\maketitle
\begin{abstract}

Conversational database interfaces face a critical challenge: users naturally embed queries in conversational noise (greetings, politeness, off-topic remarks), which degrades intent classification accuracy and wastes computational resources. Despite advances in orchestration and retrieval strategies, a fundamental question remains unanswered: which pooling strategy maximizes intent classification accuracy under realistic conversational noise in production language models? This work addresses this gap through 360 controlled experiments spanning four pooling configurations (mean, max, last-token, attention, and FFT-augmented variants) using Llama-3.2-1B-Instruct on BANKING77 and CLINC150 datasets under clean/noisy conditions with ten random seeds. Key findings reveal that attention pooling consistently outperforms alternative strategies under noisy conditions ($\approx\!+2.6$--$2.8$ F1 over the default), while mean pooling degrades performance by up to $\sim\!5$ F1 points. Frequency-domain filtering does not produce consistent accuracy improvements and functions primarily as a structural variation rather than an accuracy-enhancing component. These results provide concrete, evidence-based guidance for building noise-robust conversational classifiers: attention pooling is recommended for noisy interfaces, mean pooling should be avoided, and last-token pooling is appropriate for clean-query scenarios.

\end{abstract}

\section{Introduction}

The rise of conversational database interfaces powered by Large Language Models (LLMs) has transformed how users access structured and unstructured data. In industrial settings such as manufacturing procurement, users must retrieve part specifications, supplier availability, compliance records, pricing histories, and lead times across heterogeneous databases, where even minor retrieval errors can have significant operational and financial consequences. Rather than writing Structured Query Language (SQL) queries, users increasingly interact conversationally (e.g., ``Hey there! Can you pull up sales data from Q4? Thanks!''~\cite{Dai2024, zhu-etal-2025-plangpt-vl}).

This conversational paradigm introduces a critical challenge: separating core database intent from conversational noise, including greetings, politeness markers, discourse fillers, and off-topic remarks. In systems such as Universal Query Engine (UQE)~\cite{Dai2024} and PlanGPT~\cite{zhu-etal-2025-plangpt-vl}, failure to isolate intent leads to retrieval mismatches, misrouting across specialized models~\cite{lee-etal-2024-orchestrallm}, increased Application Programming Interface (API) costs, and degraded user experience. Accurate intent classification is therefore the foundational step for reliable conversational database systems.

Existing approaches address this challenge primarily through architectural orchestration, retrieval optimization, and confidence calibration. Retrieval-augmented generation (RAG)~\cite{lewis2020retrieval}, instruction fine-tuning~\cite{han2025alignmentcentricparadigmsurveyinstruction}, and multistep reasoning via Chain-of-Thought prompting~\cite{wei2022chain} improve downstream reasoning but do not directly resolve representation-level ambiguity introduced by conversational noise.

Despite extensive work on conversational database systems and routing strategies, a fundamental question remains insufficiently studied: how should token-level hidden states be pooled to maximize intent classification accuracy under realistic conversational noise? While mean, max, last-token, and attention-based pooling are widely used, no systematic evaluation isolates their impact in noisy, low-resource conversational LLM deployments typical of industrial settings.

This work addresses this research gap through 360 controlled experiments (10 seeds × nine experimental configurations × two datasets × two noise conditions), systematically evaluating standard pooling strategies and their variants augmented by Fast Fourier Transform (FFT) applied to transformer hidden states. Contrary to widespread practice, attention pooling consistently outperforms alternatives under conversational noise, while mean pooling degrades performance. Frequency-domain filtering yields no consistent accuracy improvements and does not enhance predictive performance relative to standard configurations.

These findings provide production-ready guidance for practitioners building noise-robust conversational database systems, particularly in industrial environments where classification accuracy directly translates to reduced API costs and fewer failed queries. Modular code is publicly available \ifanonymize at an anonymous repository\else \cite{bogdan_bogachov_2026_18598129}\fi.

The remainder of this paper is structured as follows: Section~\ref{sec:related-work} reviews related work on conversational database interfaces, pooling strategies, and frequency-domain methods. Section~\ref{sec:method} outlines the method, covering the proposed ablation configurations. Section~\ref{sec:experiments-and-results} presents experimental methodology and results. Section~\ref{sec:conclusion} provides conclusions and discusses the implications of this work.

\section{Related work}
\label{sec:related-work}

\subsection{Conversational database systems}

The proposed method targets conversational database interfaces, where natural language queries must be robustly mapped to structured operations despite conversational noise. Recent systems address this challenge through architectural and retrieval-based strategies. OrchestraLLM~\cite{lee-etal-2024-orchestrallm} routes dialogue turns between fine-tuned small language models and large language models via nearest-neighbor retrieval over expert pools, improving accuracy while reducing LLM calls, though primarily evaluated in dialogue state tracking. UQE~\cite{Dai2024} extends SQL with natural-language predicates for querying unstructured data, combining multimodal embeddings, stratified sampling, and execution planning to reduce LLM usage while maintaining accuracy. For domain specialization, PlanGPT~\cite{zhu-etal-2025-plangpt-vl} integrates domain-tuned embeddings, retrieval, instruction fine-tuning, and orchestration agents for urban planning tasks. Additionally, SQL-aware confidence estimation methods~\cite{maleki2025confidenceestimationtexttosqllarge} derive reliability scores directly from token-level probabilities.

\subsection{Pooling strategies}

Pooling aggregates token-level representations into fixed-size sequence representations for downstream classification. Common strategies include: (i) \emph{special-token pooling}, such as using BERT's [CLS] representation for sequence classification~\cite{devlin-etal-2019-bert}, (ii) \emph{last-token pooling}, commonly used with decoder-only models, (iii) \emph{mean pooling} (simple average across tokens)~\cite{reimers-gurevych-2019-sentence}, (iv) \emph{max pooling} (per-dimension maximum)~\cite{ennadir2025poolwiselyeffectpooling}, and (v) \emph{attention-based pooling} (learned weighted sum over tokens)~\cite{Chen20233963}.

While these pooling strategies are widely used, systematic comparisons remain sparse and domain-specific. BERT uses the final hidden representation of the special [CLS] token for sequence-level classification~\cite{devlin-etal-2019-bert}. Recent work reveals substantial variation across tasks: some show that mean pooling is highly effective for BERT-based sentiment classification~\cite{xing2025comparativeanalysispoolingmechanisms}, while max pooling outperforms average and attention pooling for biomedical relation extraction in Natural Language Processing (NLP) models based on Convolutional Neural Networks (CNN)~\cite{Suárez-Paniagua2018}. Recent comprehensive ablations~\cite{ennadir2025poolwiselyeffectpooling} confirm that pooling choice critically impacts transformer performance across domains; however, these analyses do not explicitly target conversational intent classification or evaluate robustness under realistic conversational noise.

\subsection{Frequency-domain methods}

Frequency-domain filtering draws motivation from spectral bias theory~\cite{rahaman2019spectral}, where neural networks prioritize low-frequency patterns. FNet~\cite{lee-thorp-etal-2022-fnet} replaced self-attention with 2D Fourier transforms, achieving 92--97\% of BERT performance with 80\% faster training. Recent work~\cite{lao2024frequency,xu2020learning,gao2024parameter} applies frequency filtering to representations, inputs, and weights, demonstrating low-frequency components capture semantics while high-frequency components often represent noise. While prior work examines frequency filtering of intermediate representations~\cite{Lin202311797} and attention-based pooling under noisy conditions~\cite{Brothers20255594}, no prior studies combine 1D FFT low-pass filtering of hidden state sequences with systematic pooling strategy ablation, particularly in low-resource conversational settings. The current study evaluates 1D FFT low-pass filtering of transformer hidden states prior to pooling, a setting that has not been systematically examined in prior frequency-based work, particularly for conversational intent classification.

In summary, this research fills the identified gaps through systematic evaluation under industrial constraints: limited labeled data, conversational noise, and deployment efficiency.

\section{Method}
\label{sec:method}

\subsection{Current chatbot architecture}

\begin{figure*}[h!]
  \centering
  \includegraphics[width=\linewidth]{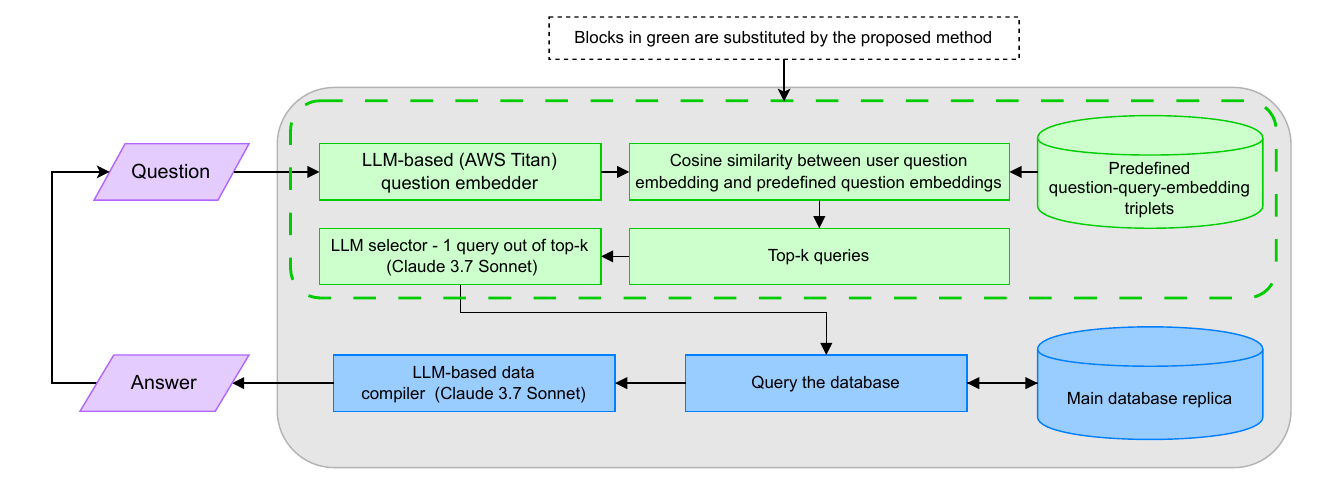}
  \caption{Overview of the chatbot pipeline. A user query is embedded and matched against predefined query templates. The selected template is used to query the database and results are compiled by an LLM to generate the final response. The blocks in green were substituted by the proposed method in Figure~\ref{fig:experiments_illustration}.}
  \label{fig:current_chatbot}
\end{figure*}

This work is done in collaboration with \ifanonymize anonymous Inc.\else Axya Inc. (Axya)\fi, an industry partner of the subject research group. \ifanonymize The anonymous partner \else Axya \fi requires a robust conversational database system capable of retrieving engineering information from their databases. Figure~\ref{fig:current_chatbot} illustrates the current architecture of the chatbot pipeline. A user question is directly embedded using an LLM-based question embedder (AWS Titan) and compared against a set of predefined question embeddings. The top-$k$ most similar predefined question-query-embedding triplets are selected, from which an LLM-based selector (Claude 3.7 Sonnet) chooses a single SQL query template. This template is then used to query a read-only replica of the main database. The retrieved results are passed to an LLM-based data compiler (Claude 3.7 Sonnet), which aggregates the information and generates the final response returned to the user.

However, the embedding computation, similarity matching against predefined queries, and subsequent selection introduce significant computational overhead; addressing this inefficiency motivates the proposed method described next.

\subsection{Proposed intent classification pipeline}

This subsection describes the lightweight intent classifier replacing the embedding+matching+selection pipeline (green blocks in Figure~\ref{fig:current_chatbot}). The core contribution is a systematic evaluation of four pooling configurations with and without FFT filtering for transformer hidden state aggregation under conversational noise. Figure~\ref{fig:experiments_illustration} illustrates the pipeline.

Hidden states $\mathbf{H} \in \mathbb{R}^{B \times T \times d}$ are extracted from the decoder of Llama-3.2-1B-Instruct~\cite{grattafiori2024llama}. These are optionally FFT-filtered, pooled into sequence representations $\mathbf{z} \in \mathbb{R}^{B \times d}$, and classified via a linear head.

\subsubsection{Pooling strategies (primary ablation)}

Four standard pooling strategies aggregate hidden states into fixed-size representations~\cite{ennadir2025poolwiselyeffectpooling, Brothers20255594}:

Let $\mathbf{H}_{i,t} \in \mathbb{R}^d$ denote hidden states, $\mathbf{z}_i \in \mathbb{R}^d$ the pooled representation, $m_{i,t} \in \{0,1\}$ the attention mask, and $T$ the sequence length, where $i \in \{1,\dots,B\}$ indexes sequences in the batch and $t \in \{1,\dots,T\}$ indexes token positions.
\begin{itemize}
\item \textbf{Mean pooling}: $\mathbf{z}_i = \frac{\sum_{t=1}^{T} m_{i,t} \mathbf{H}_{i,t}}{\sum_{t=1}^{T} m_{i,t}}$.
\item \textbf{Max pooling}: $\mathbf{z}_i = \max_{t=1}^{T} \mathbf{H}_{i,t}$, with padding masked as $-\infty$.
\item \textbf{Last-token pooling}: $\mathbf{z}_i = \mathbf{H}_{i,T_i}$, where $T_i = \sum_{t=1}^{T} m_{i,t}$.
\item \textbf{Attention pooling}: $\mathbf{z}_i = \sum_{t=1}^{T} \alpha_{i,t} \mathbf{H}_{i,t}$, where $\alpha_{i,t} = \frac{m_{i,t}\exp(\mathbf{w}^\top \mathbf{H}_{i,t})}{\sum_{s=1}^{T} m_{i,s}\exp(\mathbf{w}^\top \mathbf{H}_{i,s})}$ and $\mathbf{w} \in \mathbb{R}^d$ is learned.
\end{itemize}

\subsubsection{FFT filtering (secondary ablation variant)}

As a controlled ablation, 1D FFT low-pass filtering is optionally applied \emph{before} pooling:
\begin{equation}
  \label{eq:fft-filter}
  \tilde{\mathbf{H}} = \text{IFFT}\left( \text{FFT}(\mathbf{H}) \odot \mathbf{M}_{f_c} \right)
\end{equation}
The low-pass mask $\mathbf{M}_{f_c}$ retains the lowest $f_c = 40\%$ frequency components, determined through preliminary ablation testing cutoff ratios from 10\% to 100\%. Cutoffs below 40\% degraded performance due to excessive loss of semantic information, while higher ratios yielded no improvement over unfiltered baselines, indicating that high-frequency components contribute minimally to intent classification under conversational noise. For experimental simplicity, the cutoff was held fixed across all experiments. While adaptive or learnable spectral filtering may be promising, our objective was not to optimize frequency-domain methods, but to assess whether frequency filtering provides sufficient benefit to justify its inclusion in production intent classifiers. This setup isolates spectral noise removal as an orthogonal factor to pooling choice~\cite{Lin202311797}.

\section{Experiments and results}
\label{sec:experiments-and-results}
 
\subsection{Dataset}

Experiments are conducted on two public intent classification benchmarks: BANKING77~\cite{casanueva-etal-2020-efficient} (13{,}083 queries across 77 banking intents) and CLINC150~\cite{larson-etal-2019-evaluation} (23{,}700 queries total: 22{,}500 in-scope queries covering 150 intents across 10 domains, plus 1{,}200 out-of-scope queries). BANKING77 serves as a proxy for the manufacturing procurement domain of the industry partner, while CLINC150 provides complementary multi-domain generalization and out-of-scope detection capabilities. \ifanonymize Proprietary data could not be used due to data sharing constraints.\else Axya cannot share real customer conversations due to privacy, legal, and contractual constraints.\fi

Evaluation is performed on both raw queries and synthetically noised variants designed to emulate realistic conversational inputs. Example noise transformations are provided in Appendix~\ref{appendix-a}.

\subsection{Experimental design}

The experiments systematically disentangle the effects of pooling strategy and frequency-domain filtering under fixed training conditions. Let $\mathcal{P} = \{\text{mean}, \text{max}, \text{last}, \text{attn}\}$ denote the set of pooling operators and let $f \in \{0, 1\}$ indicate whether FFT low-pass filtering is applied. The nine experimental configurations are defined as:
\begin{equation}
  \label{eq:configs}
  \mathcal{C} = \{C_0\} \cup \{C_{p,f} : p \in \mathcal{P},\; f \in \{0, 1\}\}
\end{equation}
\noindent where $C_0$ denotes the baseline configuration using default classification head with standard last-token pooling of Llama-3.2-1B-Instruct~\cite{grattafiori2024llama}. For each remaining configuration $C_{p,f}$, the classification pipeline is:
\begin{align}
  \hat{y} = \operatorname{softmax}\left(W \cdot \operatorname{pool}_p(\mathbf{H}') + b\right) \label{eq:pipeline}
\end{align}
\noindent where $\mathbf{H}' = (1-f)\,\mathbf{H} + f\,\tilde{\mathbf{H}}$ denotes the optionally-filtered hidden states. Here, $\mathbf{H} \in \mathbb{R}^{B \times T \times d}$ are raw hidden states from the final Llama layer, $\tilde{\mathbf{H}}$ is from Eq.~\eqref{eq:fft-filter}, $\operatorname{pool}_p$ is from Section~3.2.1, and $W, b$ are linear classifier parameters. When $f=0$, this reduces to direct pooling of $\mathbf{H}$ ($C_1$--$C_4$); when $f=1$, FFT filtering precedes pooling ($C_5$--$C_8$).

This design isolates three effects:
\begin{enumerate}
    \item \textbf{Baseline ($C_0$)}: Default head performance and efficiency.
    \item \textbf{Pooling ablation ($C_1$--$C_4$)}: Effect of pooling strategy on raw hidden states.
    \item \textbf{FFT augmentation ($C_5$--$C_8$)}: Frequency filtering impact across pooling strategies.
\end{enumerate}

\begin{figure*}[h!]
  \centering
  \includegraphics[width=\linewidth]{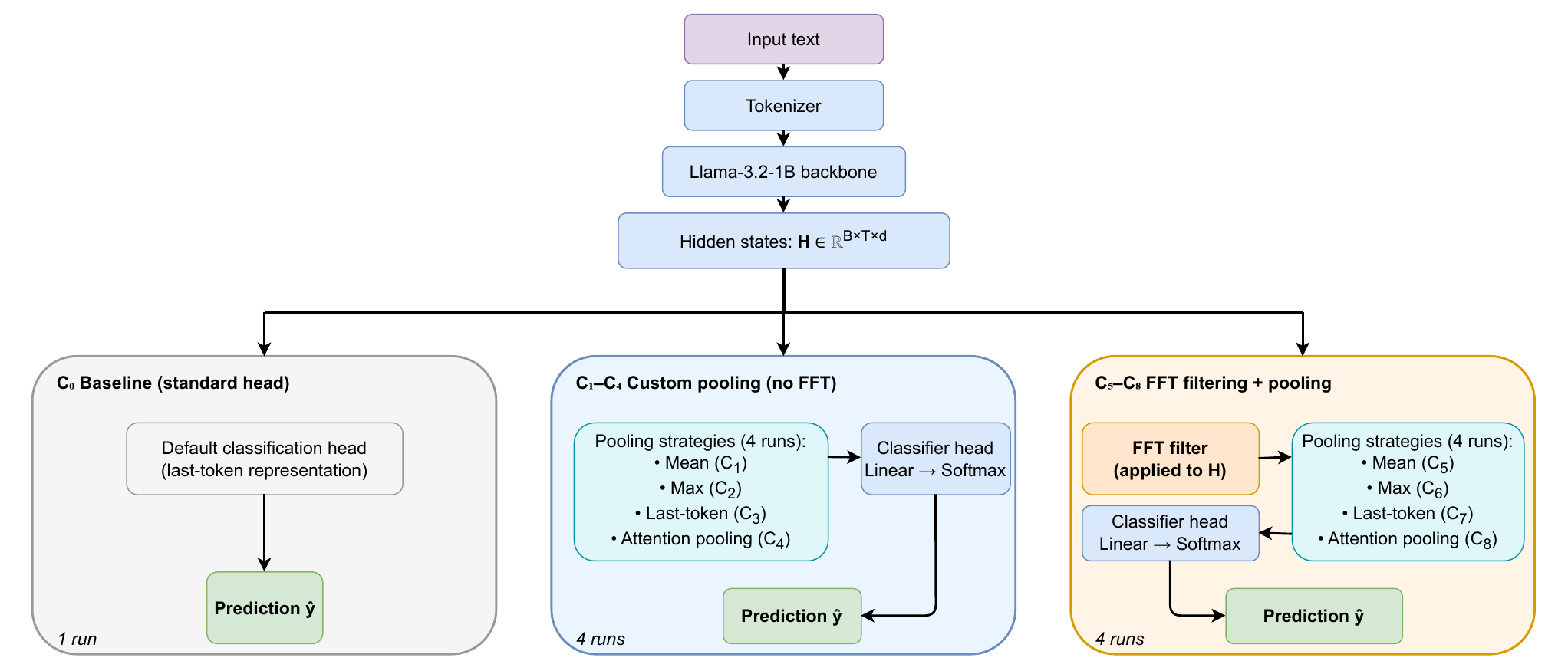}
  \caption{Overview of the nine experimental configurations. The shared backbone (input, tokenizer, Llama, $\mathbf{H}$) feeds into three groups: $C_0$ (baseline with default head), $C_1$--$C_4$ (custom pooling head applied directly to $\mathbf{H}$), and $C_5$--$C_8$ (FFT low-pass filtering applied before pooling, $\tilde{\mathbf{H}} = \text{IFFT}(\text{FFT}(\mathbf{H}) \odot \mathbf{M}_{40\%})$).}
  \label{fig:experiments_illustration}
\end{figure*}

Figure~\ref{fig:experiments_illustration} and Table~\ref{tab:experiment_design} provide visual and tabular overviews. All configurations use identical training regimes and inference pathways for fair comparison under low-resource conditions.

\begin{table}[t]
\centering
\footnotesize
\setlength{\tabcolsep}{3pt}
\begin{tabular}{clccc}
\hline
& \textbf{Configuration} & \textbf{Head} & \textbf{Pooling} & \textbf{FFT} \\
\hline
$C_0$ & Baseline & Default & Last & \xmark \\
\hline
$C_{1}$ & Mean & Custom & Mean & \xmark \\
$C_{2}$ & Max & Custom & Max & \xmark \\
$C_{3}$ & Last-token & Custom & Last & \xmark \\
$C_{4}$ & Attention & Custom & Attention & \xmark \\
\hline
$C_{5}$ & FFT + Mean & Custom & Mean & \cmark \\
$C_{6}$ & FFT + Max & Custom & Max & \cmark \\
$C_{7}$ & FFT + Last-token & Custom & Last & \cmark \\
$C_{8}$ & FFT + Attention & Custom & Attention & \cmark \\
\hline
\end{tabular}
\caption{Complete experimental design matrix. $C_0$ is the baseline using the default Llama classification head; $C_1$--$C_4$ use a custom pooling head without filtering; $C_5$--$C_8$ add FFT low-pass filtering before pooling.}
\label{tab:experiment_design}
\end{table}

Table~\ref{tab:training_config} in Appendix~\ref{appendix-b} summarizes the training and adaptation hyperparameters used across all experiments. The setup reflects a low-resource, efficiency-oriented regime, combining parameter-efficient fine-tuning with early stopping and constrained checkpointing to minimize compute and storage overhead.

Experiments were executed on Linux nodes. Each training run used a single NVIDIA H100-80GB GPU with 1 CPU core and 16GB system RAM, and multiple such single-GPU jobs were launched concurrently to run different configurations in parallel.

Ten independent experiments are conducted per configuration using fixed, predefined seeds, with results reported as mean $\pm$ standard deviation across seeds in Table~\ref{tab:complete_results}. Exact seeds are provided in Appendix~\ref{appendix-c}.

\subsection{Results}

Table~\ref{tab:complete_results} presents comprehensive results across all nine experimental configurations and four datasets (Clean/Noisy BANKING77, Clean/Noisy CLINC150), reporting F1 scores (mean ± std dev over 10 seeds) and FLOPs. All F1 scores are reported in percentage points. Figures~\ref{fig:f1_banking} and~\ref{fig:f1_clinc} in Appendix~\ref{appendix-d} visualize F1 performance for all configurations.

\begin{table*}[t]
\centering
\scriptsize
\resizebox{\textwidth}{!}{%
\begin{tabular}{lcccccccc}
\toprule
& \multicolumn{2}{c}{Clean BANKING77} & \multicolumn{2}{c}{Noisy BANKING77} & \multicolumn{2}{c}{Clean CLINC150} & \multicolumn{2}{c}{Noisy CLINC150} \\
\cmidrule(lr){2-3} \cmidrule(lr){4-5} \cmidrule(lr){6-7} \cmidrule(lr){8-9}
 & F1 (\%) & Train FLOPs & F1 (\%) & Train FLOPs & F1 (\%) & Train FLOPs & F1 (\%) & Train FLOPs \\
\midrule
default & 88.80 ± 0.98 & 2.95e17 ± 2.57e16 & 87.67 ± 0.64 & 2.91e17 ± 2.58e16 & 86.66 ± 1.15 & 3.60e17 ± 2.09e16 & 86.42 ± 1.56 & 3.73e17 ± 1.76e16 \\
\midrule
custom~mean & 88.65 ± 0.54 & 3.97e17 ± 3.26e16 & 86.43 ± 0.53 & 4.09e17 ± 1.91e16 & 83.11 ± 1.25 & 4.97e17 ± 2.14e16 & 83.66 ± 1.59 & 4.94e17 ± 4.84e16 \\
custom~max & 89.63 ± 0.39 & 3.96e17 ± 2.10e16 & 88.89 ± 0.37 & 3.83e17 ± 1.71e16 & 88.26 ± 1.16 & 4.72e17 ± 2.37e16 & 88.88 ± 0.98 & 4.94e17 ± 3.93e16 \\
custom~last & \textbf{90.79 ± 0.37} & 2.73e17 ± 1.97e16 & 89.97 ± 0.50 & 2.80e17 ± 1.96e16 & \textbf{89.56 ± 0.58} & 3.26e17 ± 1.79e16 & 88.78 ± 0.71 & 3.45e17 ± 1.02e16 \\
custom~attention & 90.74 ± 0.43 & 2.85e17 ± 2.28e16 & \textbf{90.31 ± 0.36} & 2.98e17 ± 2.02e16 & 88.81 ± 0.98 & 3.18e17 ± 1.81e16 & \textbf{89.25 ± 1.12} & 3.22e17 ± 2.11e16 \\
\midrule
custom~FFT~mean & 82.17 ± 2.30 & 3.96e17 ± 2.08e16 & 86.38 ± 0.51 & 4.05e17 ± 2.25e16 & 83.25 ± 1.55 & 5.00e17 ± 2.29e16 & 83.82 ± 1.65 & 5.03e17 ± 4.52e16 \\
custom~FFT~max & 89.07 ± 0.59 & 4.03e17 ± 1.68e16 & 88.37 ± 0.92 & 3.76e17 ± 4.10e16 & 86.84 ± 0.99 & 4.75e17 ± 2.43e16 & 87.75 ± 1.11 & 4.66e17 ± 1.92e16 \\
custom~FFT~last & 90.42 ± 0.51 & 2.78e17 ± 2.09e16 & 89.33 ± 0.62 & 2.76e17 ± 2.08e16 & 86.95 ± 1.08 & 3.19e17 ± 1.22e16 & 87.69 ± 1.00 & 3.51e17 ± 1.64e16 \\
custom~FFT~attention & 84.49 ± 0.81 & 2.76e17 ± 2.29e16 & 90.19 ± 0.35 & 2.88e17 ± 2.20e16 & 86.79 ± 1.04 & 3.23e17 ± 2.48e16 & 88.32 ± 0.89 & 3.30e17 ± 2.44e16 \\
\bottomrule
\end{tabular}%
}
\caption{Performance metrics (F1 scores and training FLOPs) across experiment configurations for each dataset.}
\label{tab:complete_results}
\end{table*}

\subsubsection{Pooling performance under clean vs. noisy conditions}

Custom pooling heads consistently outperform the default head across all datasets, with particularly strong improvements under noisy conditions.

\textbf{Clean datasets} systematically favor \emph{last-token pooling} ($C_3$: F1=90.79 on BANKING77 and 89.56 on CLINC150). This aligns with the intuition that clean conversational inputs preserve intent signals in final tokens, where discourse markers and politeness phrases minimally corrupt the representation. The inferior performance of the default head ($C_0$: F1=88.80 on BANKING77 and 86.66 on CLINC150) despite using last-token pooling stems from two factors: (1) computing logits across \emph{all} sequence positions before final-token selection introduces gradient noise during fine-tuning, and (2) suboptimal initialization of the default projection layer relative to the custom linear head trained end-to-end on pooled representations.

\textbf{Noisy datasets}, conversely, favor \emph{attention pooling} ($C_4$: F1=90.31 on BANKING77 and 89.25 CLINC150), achieving 2.64\% and 2.83\% gains over the default, respectively. The learned token weighting mechanism of attention pooling dynamically suppresses conversational noise (greetings, hedging phrases, discourse markers), prioritizing intent-carrying tokens regardless of position.

\subsubsection{FFT filtering effects}

FFT low-pass filtering yields no consistent accuracy improvements over non-FFT counterparts. On \textbf{clean datasets}, FFT degrades performance (e.g., FFT-attention $C_8$: F1=84.49 vs. attention $C_4$: 90.74 on BANKING77 and FFT-attention $C_8$: F1=86.79 vs. attention $C_4$: 88.81 on CLINC150), as the 40\% cutoff discards high-frequency components containing discriminative intent signals preserved in clean sequences. This signal loss explains the expected F1 drop, since low-pass filtering removes fine-grained token distinctions critical for precise classification.

On \textbf{noisy datasets}, FFT-attention ($C_8$) approaches non-FFT attention performance ($C_4$) ($\Delta$F1=-0.12 on BANKING77 and -0.93 on CLINC150) while achieving similar performance with marginally reduced compute in some settings. The modest F1 degradation suggests that FFT suppresses dominant noise components without substantially harming core intent representations.

FFT variants do not consistently improve accuracy over their non-FFT counterparts, and the observed differences are small relative to run-to-run variation, indicating negligible accuracy benefits overall.

\subsubsection{Efficiency characteristics}

All configurations exhibit comparable training FLOPs (2.73--5.03$\times 10^{17}$), with FFT and non-FFT variants demonstrating similar computational profiles across datasets.

\subsubsection{Robustness}

Across seeds and both datasets, the ranking is consistent (attention best under noise; last-token best on clean inputs), indicating the effects are not seed-dependent. Code is available \ifanonymize at an anonymous repository\else \cite{bogdan_bogachov_2026_18598129}\fi.
\section{Conclusion}
\label{sec:conclusion}

This study examines intent classification in conversational database interfaces under user-generated noise. Across 360 controlled experiments (nine experimental configurations, four datasets -- Clean/Noisy BANKING77 and CLINC150 -- and ten seeds), it provides evidence-based guidance for practice.

Results reveal that custom \emph{attention pooling} performs best under noisy conditions (F1 = 90.31 / 89.25 on BANKING77 / CLINC150; +2.64 / +2.83 over default), learning to suppress greetings and discourse markers while preserving intent-bearing tokens. \emph{Last-token pooling} dominates clean datasets (F1 = 90.79 / 89.56 on BANKING77 / CLINC150; +1.99 / +2.90 over default), benefiting from stable end-position intent signals absent in the default head’s logit computation.

FFT low-pass filtering yields no accuracy gains and degrades clean performance via discriminative signal loss.

The findings have practical implications for industrial applications: use attention pooling for noisy or multi-turn conversational inputs; reserve last-token pooling for short, clean queries; avoid mean pooling due to consistent degradation; apply max pooling only when intent hinges on a few highly discriminative tokens; FFT filtering can be considered as a structural alternative, with accuracy differences remaining small and computational costs comparable to standard configurations.

The conclusions are derived from public intent benchmarks with synthetic conversational noise and a constrained model setting (Llama-3.2-1B-Instruct with a fixed FFT cutoff), and may not fully capture all production noise patterns or domain-specific intent taxonomies. A detailed discussion of limitations and transfer considerations is provided in Appendix~\ref{appendix-limitations}.

These findings address a key upstream bottleneck for orchestration, retrieval, and SQL generation pipelines, all dependent on accurate first-step intent classification. For manufacturing procurement platforms and similar systems, attention pooling is recommended as the default for noisy interfaces, with last-token pooling limited to clean-query scenarios. The modular classification head integrates with parameter-efficient adaptation (e.g., LoRA), and all code is publicly available \ifanonymize anonymous repository\else \cite{bogdan_bogachov_2026_18598129}\fi, establishing a reproducible baseline for noise-robust LLM classifiers.

\section*{Acknowledgments}

\ifanonymize
Anonymous
\else
This research work is supported by grants from the Natural Sciences and Engineering Research Council of Canada and the Digital Research Alliance of Canada. This research work is supported by the Natural Sciences and Engineering Research Council of Canada Discovery Grant RGPIN-2018-05971, the McGill Engineering Doctoral Award (MEDA), and Mitacs Accelerate Grant IT46100. The authors thank the Digital Research Alliance of Canada for providing computational resources through Compute Canada. The authors also thank Oleg Miroshnichenko for inspiration and support.
\fi

\bibliography{custom}

\clearpage

\appendix

\section{Experimental details}
\label{appendix-a}

To emulate realistic conversational inputs, synthetic noise is applied to both BANKING77 and CLINC150 queries while preserving the original intent. The noise injection procedure augments queries with natural conversational artifacts, including greetings, contextual clauses, self-corrections, follow-up questions, soft urgency markers, and occasional mild typos. Distinct noise styles are applied to each dataset as well as training and test sets to induce a controlled distribution shift.

\textbf{Example noise transformations:}
\begin{itemize}
    \item \emph{``What can you do to unblock my pin?''} $\rightarrow$ \emph{``I'm receiving an error message -- What can you do to unblock my pin?''}
    \item \emph{``I accidentally made a transaction to the wrong account.''} $\rightarrow$ \emph{``I accidentally made a transaction to the wrong account. could you confirm what is happening?''}
\end{itemize}

This augmentation targets conversational realism rather than adversarial corruption.

\section{Training hyperparameters}
\label{appendix-b}
\setcounter{table}{0}
\renewcommand{\thetable}{\thesection.\arabic{table}}

\begin{table}[!htbp]
\centering
\small
\begin{tabular}{ll}
\hline
\textbf{Component} & \textbf{Setting} \\
\hline
\multicolumn{2}{l}{\emph{LoRA configuration}} \\
Rank ($r$) & 4 \\
Scaling factor ($\alpha$) & 16 \\
Target modules & $q\_proj$, $v\_proj$ \\
LoRA dropout & 0.1 \\
Bias & None \\
\hline
\multicolumn{2}{l}{\emph{Training setup}} \\
Max sequence length & 512 \\
Training epochs (max) & 45 \\
Learning rate & $2 \times 10^{-5}$ \\
LR scheduler & Cosine \\
Warmup ratio & 0.1 \\
Precision & FP16 \\
\hline
\multicolumn{2}{l}{\emph{Batching}} \\
Per-device train batch size & 48 \\
Per-device eval batch size & 48 \\
Effective batch size & 48 \\
Gradient accumulation & Auto \\
\hline
\multicolumn{2}{l}{\emph{Evaluation and early stopping}} \\
Evaluation strategy & Every 25 steps \\
Early stopping patience & 3 \\
Early stopping threshold & $1 \times 10^{-3}$ \\
Metric for selection & Validation loss \\
Load best model & Yes \\
\hline
\multicolumn{2}{l}{\emph{Checkpointing}} \\
Save strategy & Every evaluation step \\
Max checkpoints kept & 3 \\
\hline
\end{tabular}
\caption{Training hyperparameters used in all experiments. The setup emphasizes parameter efficiency, early convergence, and reduced computational overhead.}
\label{tab:training_config}
\end{table}

\section{Initialization seeds}
\label{appendix-c}

All experiments were repeated with 10 fixed random seeds to ensure reproducibility.
Seeds used: 42, 123, 456, 789, 1011, 1213, 1415, 1617, 1819, 2021.

\clearpage
\onecolumn
\section{Supplementary bar charts}
\label{appendix-d}
\setcounter{figure}{0}
\renewcommand{\thefigure}{\thesection.\arabic{figure}}

\begin{figure}[h!]
  \centering
  \includegraphics[width=0.9\textwidth]{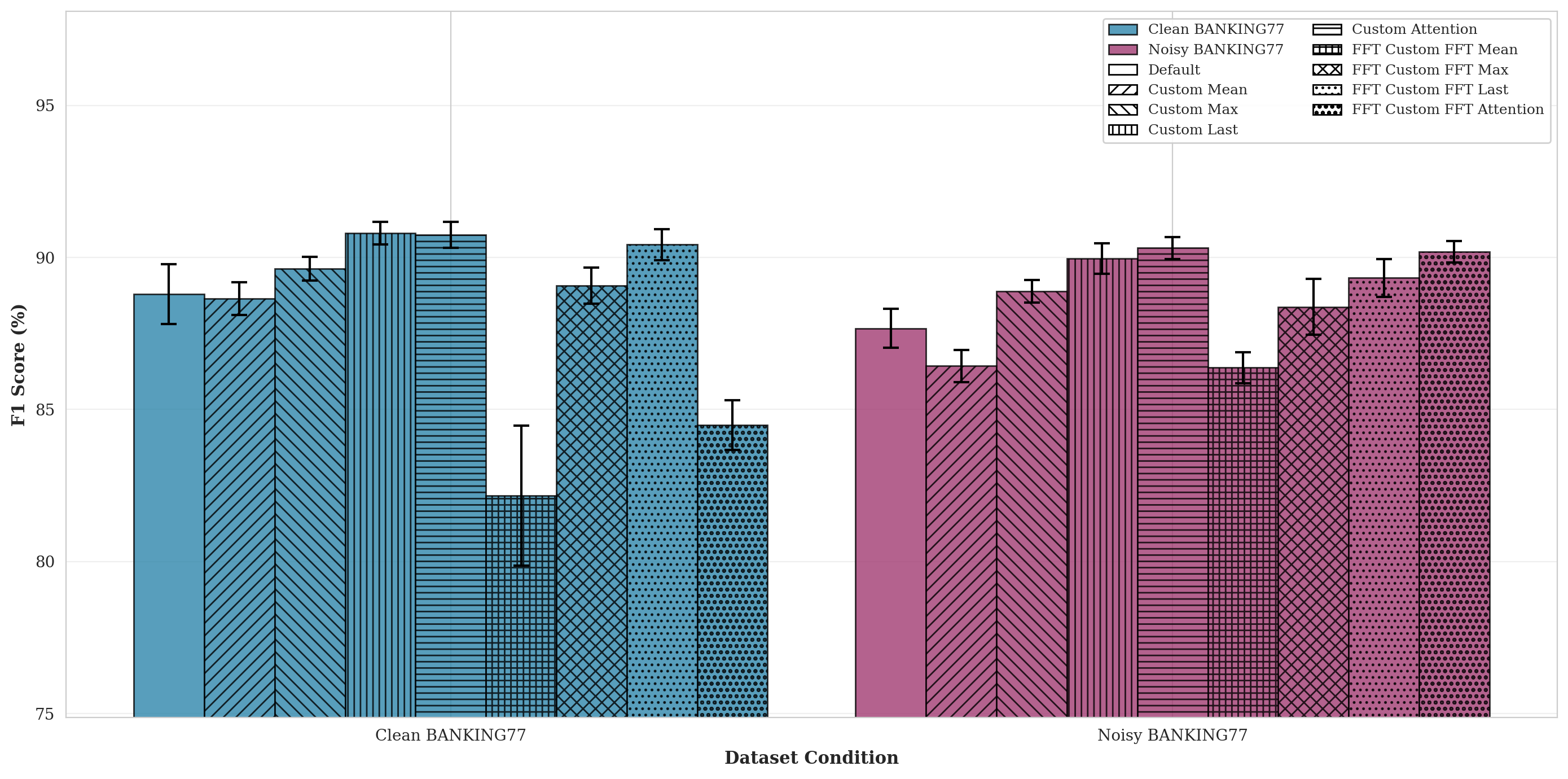}
  \caption{Mean $\pm$ standard deviation F1 scores across all experiments on the BANKING77 dataset.}
  \label{fig:f1_banking}
\end{figure}

\begin{figure}[h!]
  \centering
  \includegraphics[width=0.9\textwidth]{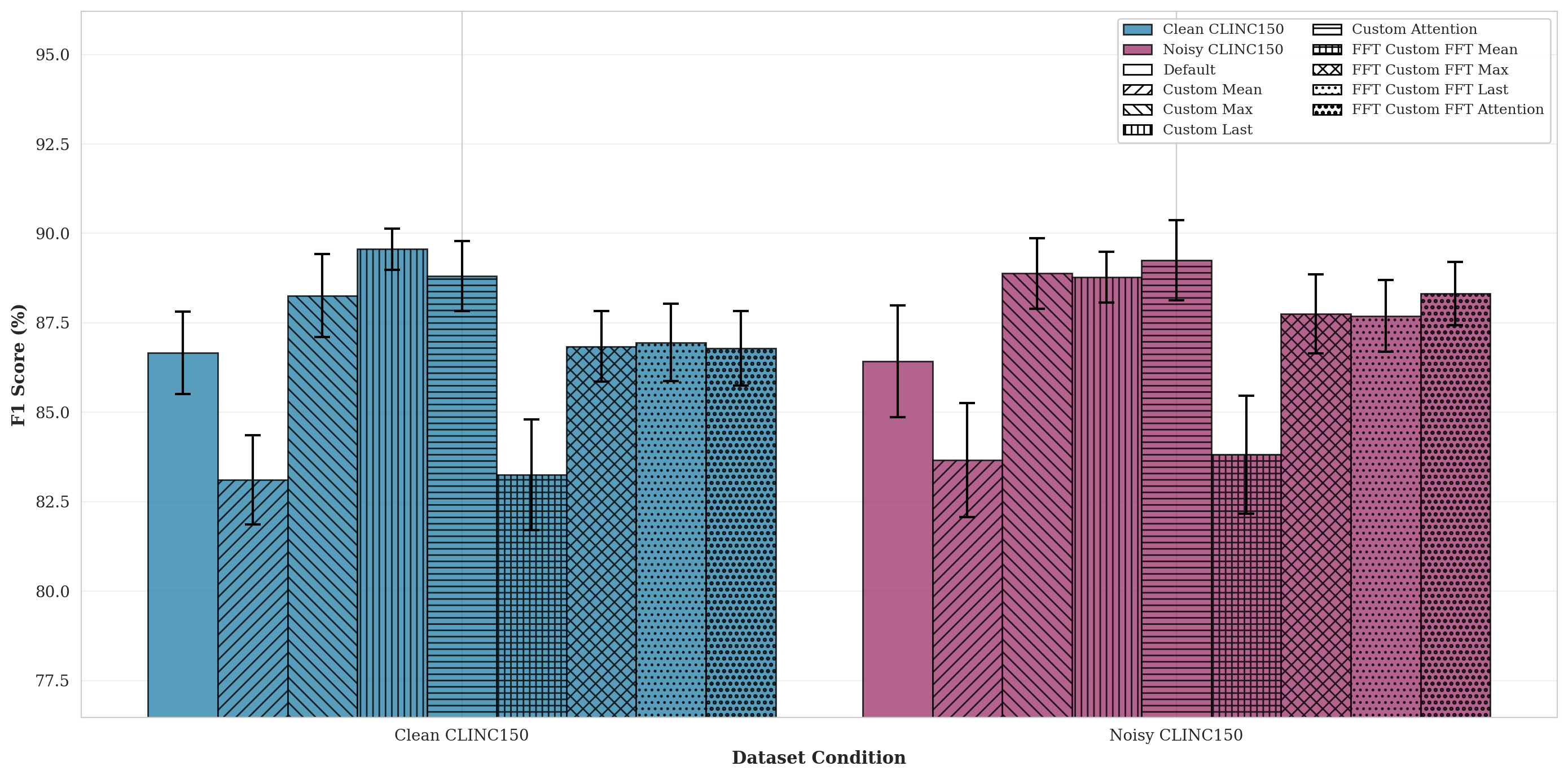}
  \caption{Mean $\pm$ standard deviation F1 scores across all experiments on the CLINC150 dataset.}
  \label{fig:f1_clinc}
\end{figure}

\clearpage
\twocolumn

\section{Limitations}
\label{appendix-limitations}

This study evaluates pooling strategies for intent classification under synthetic conversational noise in a controlled experimental setting. Although the results are consistent across 360 runs, several limitations should be noted.

The experiments rely on public benchmarks (BANKING77 and CLINC150) that serve only as proxies for industrial conversational database applications. Domain-specific intent structures, alphanumeric identifiers, and schema constraints common in real deployments are not fully represented in these datasets, so absolute performance and effect sizes may differ in production.

The noisy condition is generated via rule-based augmentation and may not fully reflect real traffic, which can include code-switching, speech-to-text errors, domain jargon, incomplete fragments, multi-intent turns, and adversarial or non-cooperative inputs. The relative ranking of pooling strategies may therefore shift under different noise distributions.

All experiments are conducted using Llama-3.2-1B-Instruct with a linear classification head and parameter-efficient fine-tuning. Results may not directly transfer to larger models, different architectures, or alternative layer-aggregation strategies, and pooling is applied only to final-layer hidden states.

FFT filtering is evaluated using a fixed 40\% low-pass cutoff selected through preliminary sweeps. This simplifies comparison but does not explore adaptive or learnable spectral approaches that could behave differently across datasets, sequence lengths, or noise types.

Finally, evaluation emphasizes offline F1 and training compute rather than end-to-end deployment metrics such as inference latency, routing accuracy within orchestration pipelines, downstream SQL correctness, user satisfaction, monitoring overhead, or operational cost. While code is provided, exact reproducibility may vary across hardware and software environments.

\end{document}